\documentclass[runningheads]{llncs}
\usepackage[T1]{fontenc}
\usepackage{graphicx,booktabs}
\usepackage{amsmath,amssymb}
\usepackage[expansion=false]{microtype}        
\usepackage{hyperref}
\usepackage{tikz}
\usetikzlibrary{arrows.meta,positioning,shapes,fit}
\usepackage{adjustbox}

\title{TTFT-Aware Graph Chain-of-Thought: Distance-Indexed Neural A* for Low-Hallucination Multi-Hop Medical Reasoning}
\titlerunning{Distance-Indexed Neural A* for GraphRAG}
\author{Bechir Dardouri
\and Ka\"{\i}s Zhioua
\and Yassine Msaddak
}
\authorrunning{B. Dardouri et al.}
\institute{Tanit Healthcare Technologies, Tunisia \\\email{kais.zhioua@tanit.ai}}

\begin{document}
\maketitle

\begin{abstract}
\begin{sloppypar}
Hallucinations and opaque reasoning remain unacceptable failure modes for clinical LLMs.
We present a production-grade GraphRAG stack that constrains answers to verifiable \emph{graph chain-of-thought} paths in a heterogeneous, $\sim$700K-node medical knowledge graph powering a fertility assistant.
The core idea is \emph{targeted navigation}: a \textbf{directed Pruned Landmark Labeling} (PLL) oracle provides exact distances for sub-milli\-second feasibility checks and simple-path enumeration, while a lightweight \textbf{AStarNet} heuristic operates strictly \emph{within} the PLL corridor to prioritize clinically plausible expansions.
We score and pack a small, diverse set of paths (CUI/semantic-type overlap, length prior, provenance priors) to condition generation, yielding compact prompts and improved \emph{Time to First Token} (TTFT).
On fertility-focused queries, the hybrid (\emph{PLL{+}AStarNet}) establishes a better latency/recall Pareto frontier than text-only RAG and single-component baselines, lowers TTFT, and reduces clinician-audited hallucinations while preserving explanation clarity.
The result is a practical recipe for explainable, low-hallucination multi-hop medical reasoning ready for real-world deployment.
\end{sloppypar}
\keywords{GraphRAG \and Knowledge Graph \and Heterogeneous Graphs \and Medical QA \and Explainable AI \and Pruned Landmark Labeling \and A* Search \and AStarNet \and TTFT}
\end{abstract}

\section{Introduction}
\label{sec:intro}
Large language models (LLMs) can draft clinically fluent guidance, yet ungrounded generation exposes patients and clinicians to \emph{hallucinations} and opaque reasoning.
In safety-critical settings, answers must be (i)~\emph{fact-bounded} by verifiable sources, (ii)~\emph{auditable} end-to-end, and (iii)~\emph{fast} to first token for interactive use.
We take a structure-first stance: constrain every non-trivial claim to explicit, typed relations in a heterogeneous directed medical knowledge graph (KG), yielding a verifiable \emph{graph chain-of-thought} rather than an uninspectable latent trace.

\paragraph{Why this is hard.}
The core systems obstacle is \emph{multi-hop retrieval at scale}. Beyond four hops, naïve (bi-)BFS over large typed graphs explodes in candidate paths, inflating memory and tail latency; even classical $k$-path algorithms struggle when looplessness, type constraints, and near-tie neighborhoods collide. Meanwhile, user-perceived responsiveness is dominated by \emph{Time to First Token} (TTFT), making both compute discipline and prompt compactness first-class constraints \cite{NVIDIA2024NIMTTFT,NVIDIA2024NIMParams,Databricks2023TTFT}. Text-only RAG underperforms on corpus-level and multi-hop questions; graph-structured retrieval (\emph{GraphRAG}) improves auditability but still requires efficient, targeted navigation of heterogeneous graphs \cite{Edge2024GraphRAG}.

\paragraph{Our idea.}
We convert blind enumeration into \emph{targeted navigation} by pairing an \textbf{exact distance oracle} with a \textbf{learned heuristic}. First, we adapt \emph{directed Pruned Landmark Labeling} (PLL) to answer shortest-path distances via label intersections, enabling sub-millisecond feasibility checks and meet-in-the-middle enumeration of \emph{simple} paths \cite{Akiba2013PLL,Akiba2014DynamicPLL}. Second, we serve a compact, GPU-batched \emph{AStarNet} priority strictly \emph{within} the PLL corridor to focus expansions on clinically plausible continuations \cite{Zhu2024AStarNet}. Entities are grounded via clinical NER+linking to UMLS \cite{Bodenreider2004UMLS} using \texttt{scispaCy} \cite{Neumann2019SciSpacy,scispaCyEntityLinkerDocs} and, where applicable, MedCAT \cite{Kraljevic2021MedCAT}. A lightweight scorer (CUI/semantic-type overlap, length prior, provenance priors) then selects and \emph{packs} a diverse top-$k$ of evidence paths to keep prompts compact and TTFT low while preserving auditability.

\paragraph{Contributions.}
\begin{enumerate}
  \item \textbf{Distance-indexed retrieval.} Directed PLL as an exact feasibility oracle for distance-bounded simple-path search on large medical KGs \cite{Akiba2013PLL,Akiba2014DynamicPLL}.
  \item \textbf{Neural guidance inside exact bounds.} AStarNet priorities operate within the PLL corridor, cutting expansions without sacrificing correctness \cite{Zhu2024AStarNet}.
  \item \textbf{Production-grade grounding.} TTFT-aware evidence packing and mandatory path citation reduce hallucinations and severe errors in clinician audits \cite{Asgari2025ClinicalSafety}.
\end{enumerate}

This recipe is deployed in a fertility assistant backed by a $\sim$700K-node, bimonthly-updated heterogeneous KG; on fertility queries it achieves a superior latency/recall Pareto frontier, lower TTFT, and fewer hallucinations than text-only RAG and single-component baselines.

\section{Background \& Problem}
\label{sec:bg}

\subsection{Problem definition}
\paragraph{Setup.}
We operate on a large, heterogeneous medical knowledge graph (KG) $G=(V,E,\mathcal{R})$ with directed, typed edges $e=(u \xrightarrow{r} v)$, where $r\!\in\!\mathcal{R}$ and $|V|\!\approx\!7{\times}10^{5}$. User queries are mapped to seed entities $S\!\subseteq\!V$ via clinical NER+linking to UMLS (e.g., with \texttt{scispaCy}/MedCAT) \cite{Bodenreider2004UMLS,Neumann2019SciSpacy,scispaCyEntityLinkerDocs,Kraljevic2021MedCAT}. For ordered pairs $(s,t)\!\in\!S\times S$, we must retrieve \emph{simple} (loopless) evidence paths $p:s\leadsto t$ that respect relation/type constraints and serve them as auditable conditioning for generation.

\paragraph{Assumptions and constraints.}
We assume a vetted, multi-relational KG with explicit provenance (e.g., \emph{PrimeKG}) \cite{Chandak2023PrimeKG}, periodic snapshots, and a streaming LLM that must cite graph evidence. Retrieval is bounded by strict production budgets: hop cap $H$, at most $k$ returned paths, per-pair wall time $\le T_{\max}$, memory $\le M_{\max}$, and prompt token limits that drive \emph{Time to First Token} (TTFT) \cite{NVIDIA2024NIMTTFT,NVIDIA2024NIMParams,Databricks2023TTFT}.

\paragraph{Research questions.}
\textbf{RQ1:} How can we retrieve a \emph{small}, \emph{diverse}, and \emph{auditable} set of multi-hop simple paths with high recall under tight latency/memory budgets?  
\textbf{RQ2:} Can we combine exact feasibility guarantees with learned guidance so we explore far fewer candidates without sacrificing correctness?  
\textbf{RQ3:} How should evidence be \emph{packed} to minimize prefill cost (hence TTFT) while preserving clinical auditability?

\paragraph{Why this is challenging.}
(\emph{i})~\textbf{Combinatorics}: simple-path enumeration beyond 3--4 hops explodes on typed directed graphs; naive BFS creates minute-scale tails \cite{Yen1971Loopless,Eppstein1999KShortest,Hershberger2003KSimple}.
(\emph{ii})~\textbf{Heterogeneity}: type constraints and high-degree hubs induce brittle branching factors and near-tie neighborhoods.
(\emph{iii})~\textbf{Latency}: TTFT is dominated upstream by retrieval fan-out and prompt bulk \cite{NVIDIA2024NIMTTFT,Databricks2023TTFT}.
(\emph{iv})~\textbf{Auditability}: evidence must be explicit, provenance-carrying, and cited per claim \cite{Asgari2025ClinicalSafety}.

\subsection{Existing approaches and their limitations}
We group prior solutions by core \emph{intuition}.

\paragraph{A. Text-centric RAG.}
Index unstructured corpora and condition the LLM on retrieved passages. Strengths: broad coverage and simplicity. Limitations: weak structural grounding for multi-hop reasoning, longer prompts, and higher hallucination risk; audit trails are indirect. Graph-structured retrieval (GraphRAG) improves inspectability by retrieving entities/communities/paths but still needs efficient multi-hop navigation at scale \cite{Edge2024GraphRAG}.

\paragraph{B. Brute-force or bounded graph search.}
BFS/bi-BFS with type filters; $k$-shortest(-simple) methods such as Yen/Eppstein \cite{Yen1971Loopless,Eppstein1999KShortest,Hershberger2003KSimple}. Principled but prone to memory and latency blow-ups on heterogeneous KGs; poor TTFT under realistic budgets.

\paragraph{B$'$. Point-to-point heuristic search.}
A* \cite{Hart1968AStar}, ALT landmarks \cite{Goldberg2005ALT}, and Hub Labeling \cite{Abraham2012HubLabeling} accelerate \emph{distance} queries on road/web graphs. Strength: strong P2P latency. Limitations: usually optimize \emph{one} shortest path, not a \emph{diverse top-$k$ simple} set; do not address prompt or TTFT constraints.

\paragraph{C. Label oracles and neural reasoning.}
PLL answers exact shortest-path distances via compact labels and fast intersections, with sub-ms queries at million scale \cite{Akiba2013PLL,Akiba2014DynamicPLL}; its limitation is that it returns distances, not paths. Learned policies (e.g., A*Net) prioritize local moves to reduce expansions \cite{Zhu2024AStarNet}, but without hard feasibility bounds they can drift; most focus on link prediction rather than a small, diverse, TTFT-aware auditable path set.

\paragraph{Gap.}
No prior category simultaneously provides: (\emph{i})~\emph{exact} feasibility checks to tightly bound search; (\emph{ii})~\emph{neural} guidance \emph{inside} that bound; (\emph{iii})~TTFT-aware evidence packing with mandatory citation.

\subsection{Our perspective}
\paragraph{Main insight.}
Bound what is \emph{possible} with an \textbf{exact, directed} PLL distance oracle, then focus on what is \emph{promising} using a \textbf{lightweight learned priority} (A*Net) \emph{restricted to the PLL corridor}. Finally, \textbf{pack} a small, diverse set of CUI–relation tuples and require path citation at decode time. This converts blind enumeration into \emph{targeted navigation} that respects TTFT and safety constraints while preserving recall and auditability.

\paragraph{Operational objective.}
Formally, for each $(s,t)$ let $D=d(s,t)$ from directed PLL. Enumerate only prefixes $s{\to}\cdots{\to}x$ satisfying
\begin{equation}
g(x)+d(x,t)\ \le\ D \quad \text{with node uniqueness and type masks},
\end{equation}
rank feasible frontiers via a GPU-batched learned priority, select a diverse top-$k$ under fixed time/memory/prompt budgets, and generate with mandatory path citation. \S\ref{sec:method} details this design.

\section{Method: Distance-Indexed, Neural-Guided Retrieval}
\label{sec:method}

\subsection{Scope, assumptions, and non-goals}
\paragraph{Scope.}
Given seed entities $S$ from a heterogeneous, directed medical KG $G=(V,E,\mathcal{R})$, we aim to retrieve, under strict latency/memory/prompt budgets, a \emph{small, diverse, auditable} set of \emph{simple} (loopless) multi-hop paths connecting ordered pairs $(s,t)\!\in\!S{\times}S$. The returned paths must (i) obey type/relation constraints (clinical plausibility), (ii) carry provenance, and (iii) be serialized for prompt-efficient conditioning of a streaming LLM that \emph{must} cite paths per claim.

\paragraph{Assumptions.}
We assume a vetted, multi-relational KG with periodic snapshots and explicit provenance; clinical NER with UMLS linking for seed generation; and strict production budgets $(H,k,T_{\max},M_{\max})$ together with TTFT sensitivity \cite{NVIDIA2024NIMTTFT,NVIDIA2024NIMParams,Databricks2023TTFT}.

\paragraph{Non-goals.}
We do not solve open-domain text retrieval; we do not learn the KG; and we do not modify the generator architecture beyond \emph{citation-enforcing} decoding.

\subsection{Design overview and rationale}
\begin{figure}[t]
  \centering
  \begin{adjustbox}{max width=\linewidth}
  \begin{tikzpicture}[
      >=Latex,
      font=\footnotesize
    ]
    \tikzset{
      blk/.style={
        rectangle, rounded corners=2mm, draw,
        align=center, inner sep=3pt,
        minimum height=13mm,
        text width=32mm
      },
      edgelbl/.style={
        font=\scriptsize, align=center,
        fill=white, inner sep=1pt
      },
      boxlabel/.style={
        font=\scriptsize\itshape, fill=white, inner sep=1pt
      }
    }

    \node[blk]                         (ner)  {Clinical NER\\\& Linking\\(UMLS)};
    \node[blk, right=15mm of ner]      (pll)  {Directed PLL\\Distance Oracle\\$d(s,t)$};
    \node[blk, right=15mm of pll]      (feas) {Feasible Corridor\\$g(x){+}d(x,t)\le D$\\Simple-path \& Type masks};
    \node[blk, right=15mm of feas]     (astar){A*Net Priority\\GPU-batched\\Top-$\rho$ keep};
    \node[blk, right=15mm of astar]    (score){Score \& Diversify\\Eq.~(3)\\MMR};
    \node[blk, right=15mm of score]    (pack) {ID-centric Packing\\Name Map};
    \node[blk, right=15mm of pack]     (gen)  {LLM Decode\\\& Validator\\};

    \draw[->] (ner)  -- node[edgelbl, sloped, above]{Seeds $S$}    (pll);
    \draw[->] (pll)  -- node[edgelbl, sloped, above]{$D=d(s,t)$}   (feas);
    \draw[->] (feas) --                                        (astar);
    \draw[->] (astar) -- node[edgelbl, sloped, above]{Feasible expansions} (score);
    \draw[->] (score) -- node[edgelbl, sloped, above]{Top-$k$ paths}       (pack);
    \draw[->] (pack) -- node[edgelbl, sloped, above]{Compact evidence}     (gen);

    \node[draw, dashed, rounded corners=2mm, inner sep=2mm, fit=(pll)(feas)] (box1) {};
    \node[boxlabel, above=1.2mm of box1] {Exact feasibility};

    \node[draw, dashed, rounded corners=2mm, inner sep=2mm, fit=(astar)] (box2) {};
    \node[boxlabel, above=1.2mm of box2] {Neural guidance};

    \node[draw, dashed, rounded corners=2mm, inner sep=2mm, fit=(score)(pack)] (box3) {};
    \node[boxlabel, above=1.2mm of box3] {TTFT-aware evidence};
  \end{tikzpicture}
  \end{adjustbox}
  \caption{End-to-end pipeline: PLL bounds search to the shortest-distance corridor; A*Net focuses expansions within that corridor; scoring/diversity/packing yield compact, auditable evidence for citation-enforced decoding.}
  \label{fig:pipeline}
\end{figure}
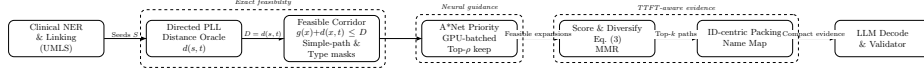

We convert blind enumeration into \emph{targeted navigation}, combining an \textbf{exact distance oracle} (directed PLL) with a \textbf{learned heuristic} (A*Net), and finishing with \textbf{TTFT-aware packing} plus \textbf{citation-enforcing decoding}. Figure~\ref{fig:pipeline} summarizes the end-to-end flow.

\paragraph{High-level loop.}
(1) Ground the prompt to seed CUIs. (2) For each $(s,t)$, query $D=d(s,t)$ from a \emph{directed} PLL service. (3) Enumerate only PLL-feasible prefixes $s{\to}\cdots{\to}x$ satisfying Eq.~\eqref{eq:feasibility}. (4) Inside that corridor, prioritize feasible expansions with GPU-batched \emph{A*Net}. (5) Score, \emph{diversify}, and \emph{pack} a top-$k$ of paths. (6) Generate with \emph{mandatory} path citation; a validator rejects out-of-evidence claims \cite{Asgari2025ClinicalSafety}.

\paragraph{Why these components (insight).}
\emph{Exactness} from PLL tightly bounds what is \emph{possible} at sub-ms cost \cite{Akiba2013PLL,Akiba2014DynamicPLL}; \emph{learned priorities} focus compute on what is \emph{promising} \cite{Zhu2024AStarNet}. This pairing collapses fan-out without sacrificing correctness, while compact packing reduces prefill tokens and TTFT.

\subsection{Component~I: Entity grounding (seeds)}
\label{subsec:seeding}
\paragraph{Procedure.}
Run clinical NER+linking (e.g., \texttt{scispaCy}, MedCAT) to map mentions to UMLS CUIs; de-duplicate, apply type allow-lists and confidence floors.

\paragraph{Justification.}
Accurate seeds bound the search space and reduce spurious fan-out; CUI normalization yields clean joins across heterogeneous edges \cite{Bodenreider2004UMLS,Neumann2019SciSpacy,Kraljevic2021MedCAT}.

\subsection{Component~II: Exact feasibility via directed PLL}
\label{subsec:pll}
\paragraph{Query.}
For each node $v$, store out-labels $L^{\text{out}}(v)$ and in-labels $L^{\text{in}}(v)$ (on $G^\top$). The exact directed distance
\[
D=\min_{w\in L^{\mathrm{out}}(s)\cap L^{\mathrm{in}}(t)} \big(d^{\mathrm{out}}(s,w)+d^{\mathrm{in}}(w,t)\big)
\]
is computed in $O(|L^{\mathrm{out}}(s)|{+}|L^{\mathrm{in}}(t)|)$ and is sub-ms in practice \cite{Akiba2013PLL}.

\paragraph{Build \& serve.}
Construct labels by pruned BFS/SSSP in a pivot order; when a visit is covered by existing labels, prune; build on $G$ and $G^\top$; serve labels from RAM behind a low-latency RPC; snapshot rebuilds \cite{Akiba2013PLL,Akiba2014DynamicPLL,PLLGitHub}.

\paragraph{Why PLL.}
It certifies feasibility \emph{exactly} at negligible latency, defining a tight corridor for path search—something heuristic-only methods cannot guarantee.

\subsection{Component~III: Distance-disciplined enumeration}
\label{subsec:feasible-search}
\paragraph{Constraint.}
Keep only prefixes $s{\to}\cdots{\to}x$ satisfying
\begin{equation}
\label{eq:feasibility}
g(x)+d(x,t)\ \le\ D,
\end{equation}
with node uniqueness (simple paths) and type/relation masks. Optionally run meet-in-the-middle: symmetric feasibility from $t$ and join frontiers.

\paragraph{Why corridor search.}
Eq.~\eqref{eq:feasibility} collapses fan-out to a predictable set while preserving all shortest-distance completions; type masks reduce low-value expansions.

\subsection{Component~IV: Neural guidance inside the PLL corridor}
\label{subsec:astarnet}
\paragraph{Priority.}
Learn $h_\theta(x\mid t,\mathcal{C})$ (context: prefix length, last relation, semantic types). Rank feasible candidates by
\[
\text{A*: } f(x)=g(x)+\lambda h_{\theta}(x\mid t,\mathcal{C})
\quad\text{or}\quad
\text{Beam: } s(x)=\lambda h_{\theta}(x\mid t,\mathcal{C})-\mu g(x),
\]
and expand under budgets (beam $B$, hop cap $H$, time/memory caps). This follows A* \cite{Hart1968AStar} but remains \emph{strictly} inside the exact PLL corridor.

\paragraph{Training.}
Shallow MLP over entity/relation embeddings, semantic-type indicators, prefix features, and $e(t)$; train on KG triples with uniform+degree-aware negatives (logistic pairwise loss); \emph{upweight} edges on short PLL-feasible prefixes to correlate with \emph{path} utility; temperature-calibrate. Serve with dynamic microbatching and keep-ratio $\rho$ (retain top-$\rho$ feasible per step) \cite{Zhu2024AStarNet,AStarNetGitHub}.

\paragraph{Why learned guidance.}
Within the exact corridor, admissibility is unnecessary; the heuristic simply prioritizes promising moves, dramatically reducing expansions with no loss of correctness.

\subsection{Component~V: Path scoring, diversity, and packing}
\label{subsec:scoring}
\paragraph{Utility.}
For $p=v_0 \xrightarrow{r_0}\!\cdots\xrightarrow{r_{\ell-1}} v_\ell$:
\begin{multline}
\label{eq:score}
\mathrm{Score}(p)=
\alpha\,\mathrm{CUIOverlap}(p)+
\beta\,\mathrm{SemTypeOverlap}(p)+
\gamma\,\tfrac{1}{1+\ell(p)} \\
{}+\delta\,\mathrm{EdgeReliability}(p)-
\eta\,\mathrm{HubPenalty}(p).
\end{multline}
Tune $(\alpha,\beta,\gamma,\delta,\eta)$ on validation for path nDCG and clinician faithfulness.

\paragraph{Diversity.}
Select $\mathcal{P}_k$ via MMR: $\max_p \lambda\,\mathrm{Score}(p)-(1-\lambda)\max_{q\in\mathcal{P}} \mathrm{Sim}(p,q)$, with Jaccard similarity over CUIs/relations.

\paragraph{Packing (TTFT-aware).}
Serialize as compact CUI–relation tuples with a single \texttt{Name Map} per prompt to minimize tokens; decoding \emph{requires} citing path IDs; a validator rejects out-of-evidence claims or missing citations and can trigger abstention \cite{Asgari2025ClinicalSafety}.

\paragraph{Why this head.}
Clinically shaped priors (+ hub penalty) prefer concise, plausible chains; MMR avoids near-duplicates; ID-centric packing reduces prefill and TTFT.

\subsection{Complexity and systems considerations}
\paragraph{Asymptotics.}
PLL query $O(|L^{\text{out}}(s)|{+}|L^{\text{in}}(t)|)$; enumeration is bounded by the feasible frontier induced by Eq.~\eqref{eq:feasibility}; A*Net adds negligible GPU latency due to microbatching and top-$\rho$ filtering.

\paragraph{Caching and fallbacks.}
LRU cache hot $(s,t)$ distance queries; degrade to PLL-only when GPU is cold/saturated; cap $B$ and $\rho$ under bursty load; enforce per-pair $T_{\max}$ to protect p95.

\paragraph{Alternatives considered (why not).}
\emph{ALT/Hub Labeling only} \cite{Goldberg2005ALT,Abraham2012HubLabeling}: great for \emph{distances}, but do not curate an auditable top-$k$ of simple paths or address TTFT.  
\emph{$k$-shortest(-simple) paths only} \cite{Yen1971Loopless,Eppstein1999KShortest,Hershberger2003KSimple}: hard to keep loopless/type-constrained at scale; poor tails.  
\emph{Heuristic-only search}: no exact feasibility, unstable tails, and recall loss.

\subsection{Operational algorithm (concise pseudo-code)}
\noindent\textbf{Input:} seed set $S$, hop cap $H$, budgets $(k,T_{\max},M_{\max})$ \\
\textbf{For each} ordered $(s,t)\!\in\!S{\times}S$: \\
\quad $\bullet$ Query $D\leftarrow d(s,t)$ from directed PLL. \\
\quad $\bullet$ Initialize frontier with $s$; enforce node-uniqueness and type masks. \\
\quad $\bullet$ While time/memory budget remains: \\
\qquad -- Keep only feasible nodes: $g(x)+d(x,t)\le D$. \\
\qquad -- Rank feasible nodes via $f(x)$ or $s(x)$; expand top-$\rho$ under beam $B$. \\
\qquad -- When $t$ is reached with $\ell(p)\le H$, add $p$ to candidate set. \\
\quad $\bullet$ Score candidates by Eq.~\eqref{eq:score}; select $\mathcal{P}_k$ via MMR; pack as tuples with a single \texttt{Name Map}. \\
\textbf{Output:} compact, diverse $\mathcal{P}_k$ with path IDs for citation-enforcing decoding.

\section{Experimental Setup}
\label{sec:setup}
We evaluate under production-like conditions for a fertility assistant.

\paragraph{Graph \& queries.}
\textbf{KG:} Directed, typed medical graph ($\sim$700K nodes) spanning drug--disease, phenotype, procedure, biomarker, and guideline edges; \emph{bi-monthly} snapshots.
\textbf{Seeds:} Prompts mapped to UMLS CUIs via NER+linking with \texttt{scispaCy}/MedCAT \cite{Bodenreider2004UMLS,Neumann2019SciSpacy,Kraljevic2021MedCAT}.
\textbf{Splits:} De-identified live prompts and clinician scenarios (etiology $\to$ diagnostic $\to$ intervention $\to$ outcome), stratified by PLL distance; 20\,\% val, 20\,\% test.

\paragraph{Systems \& hardware.}
\textbf{Storage:} Neo4j cluster \cite{Neo4jDocs}.
\textbf{Distance oracle:} Directed PLL service (labels in RAM), blue/green rebuild per snapshot, LRU hot-pair cache \cite{Akiba2013PLL,PLLGitHub}.
\textbf{Heuristic:} AStarNet GPU microservice with dynamic microbatching and warm pools \cite{Zhu2024AStarNet}.
\textbf{Generation:} Streaming LLM conditioned strictly on selected path IDs.
\textbf{Compute:} DGX-class system with multiple A100 GPUs \cite{NVIDIA2020DGXA100}.

\paragraph{Methods.}
We compare six configurations under identical $H$, time/memory, and evidence-token budgets:
\textbf{Text~RAG} (dense retrieval, no graph); \textbf{BFS} (depth-$H$, type-filtered); \textbf{Bi-BFS} (meet-in-the-middle);
\textbf{PLL-only} (distance-bounded enumeration); \textbf{AStarNet-only} (A*/beam, no PLL corridor);\allowbreak{} \textbf{Ours} (\emph{PLL{+}AStarNet}).
Figs.~\ref{fig:frontier-latency}--\ref{fig:frontier-ttft} and Table~\ref{tab:faithfulness} cover the three strongest baselines; Fig.~\ref{fig:frontier-expansions} adds BFS/Bi-BFS to isolate unguided traversal cost.

\paragraph{Metrics \& protocol.}
\textbf{Retrieval:} recall@$k$ vs.\ a high-budget oracle; path nDCG; cumulative expansions; \%PLL-feasible.
\textbf{Latency:} end-to-end p50/p95 and \textbf{TTFT} \cite{NVIDIA2024NIMTTFT,NVIDIA2024NIMParams,Databricks2023TTFT}.
\textbf{Memory:} peak RSS (enumerator) and peak GPU (heuristic).
\textbf{Faithfulness:} clinician audit---hallucination \%, severe error \%, explanation clarity (1--5); automatic citation-fidelity check \cite{Asgari2025ClinicalSafety}.
\textbf{Stats:} 95\% CIs via bootstrap; paired Wilcoxon for ours vs.\ baselines.

\section{Results}
\label{sec:results}
We report end-to-end performance under identical budgets for \textbf{BFS}, \textbf{Bi-BFS}, \textbf{PLL-only}, \textbf{AStarNet-only}, and \textbf{Ours} (\emph{PLL{+}AStarNet}). Metrics follow \S\ref{sec:setup}: recall@$k$, p50/p95 latency, \textbf{TTFT}, cumulative expansions, memory, and clinician-audited faithfulness. All results are on the held-out test split; CIs (95\%) are from bootstrap over prompts.

\subsection{Main Comparison: Pareto \& TTFT}
\begin{figure}[htbp]
  \centering
  \includegraphics[width=.5\linewidth]{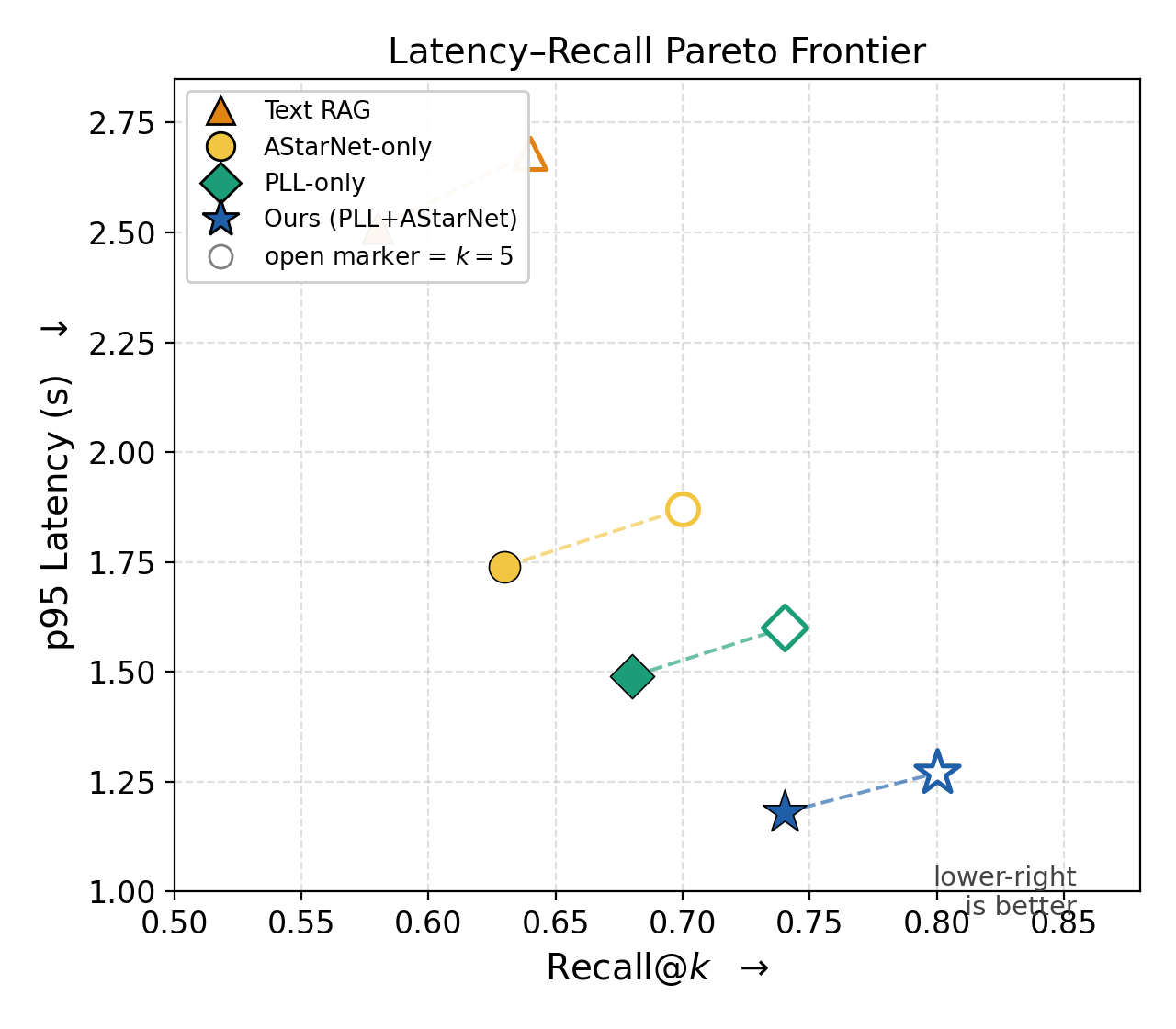}
  \caption{Latency–recall frontier (lower-right is better). Hybrid (\emph{PLL{+}AStarNet}) improves the Pareto frontier relative to Text RAG, AStarNet-only, and PLL-only at $k\!\in\!\{3,5\}$.}
  \label{fig:frontier-latency}
\end{figure}

\paragraph{Latency/recall frontier.}
Figure~\ref{fig:frontier-latency} shows recall@$k$ vs.\ p95 latency for $k\!\in\!\{3,5\}$. The hybrid establishes a new frontier: at $k\!=\!3$ it attains \textbf{0.74} recall with \textbf{1.18\,s} p95 latency, outperforming Text RAG (0.58, 2.51\,s), AStarNet-only (0.63, 1.74\,s), and PLL-only (0.68, 1.49\,s). At iso-recall 0.68, the hybrid reduces p95 latency by \textbf{21.0\%} vs.\ PLL-only and \textbf{32.2\%} vs.\ AStarNet-only.

\begin{figure}[htbp]
  \centering
  \includegraphics[width=.5\linewidth]{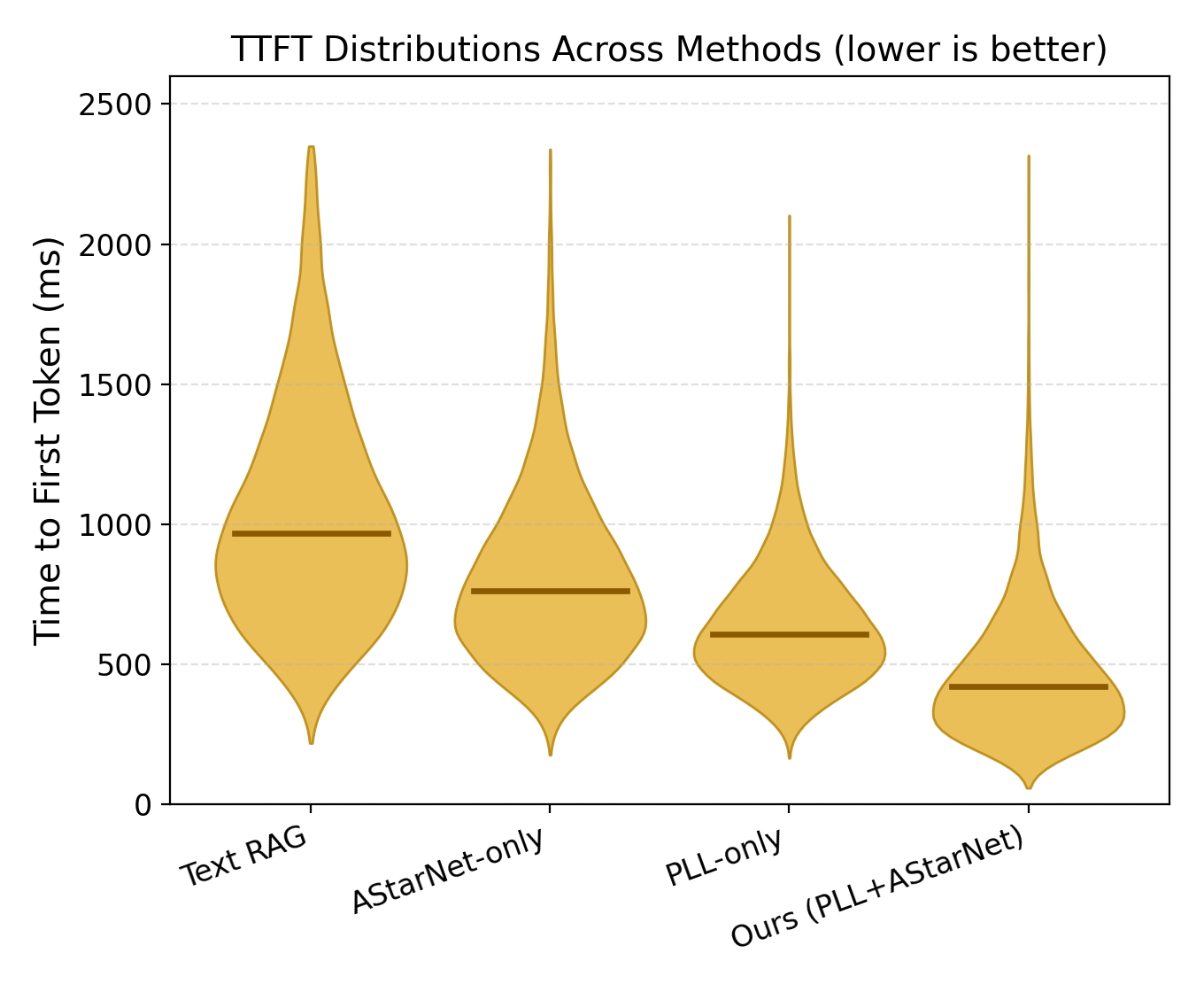}
  \caption{TTFT distributions across methods (lower is better). The hybrid lowers median TTFT and tightens the tail (p95) vs.\ all baselines.}
  \label{fig:frontier-ttft}
\end{figure}

\paragraph{TTFT distributions.}
Figure~\ref{fig:frontier-ttft} plots TTFT violins. Median TTFT drops from \textbf{980\,ms} (Text RAG) and \textbf{610\,ms} (PLL-only) to \textbf{420\,ms} (Ours), with tail (p95) shrinking from 1.92\,s to 0.93\,s. Two factors drive the shift: (i) sub-ms \emph{distance} checks gate enumeration early; (ii) ID-centric path packing cuts prefill tokens. Variance also narrows, yielding more predictable interactivity.

\begin{figure}[htbp]
  \centering
  \includegraphics[width=.5\linewidth]{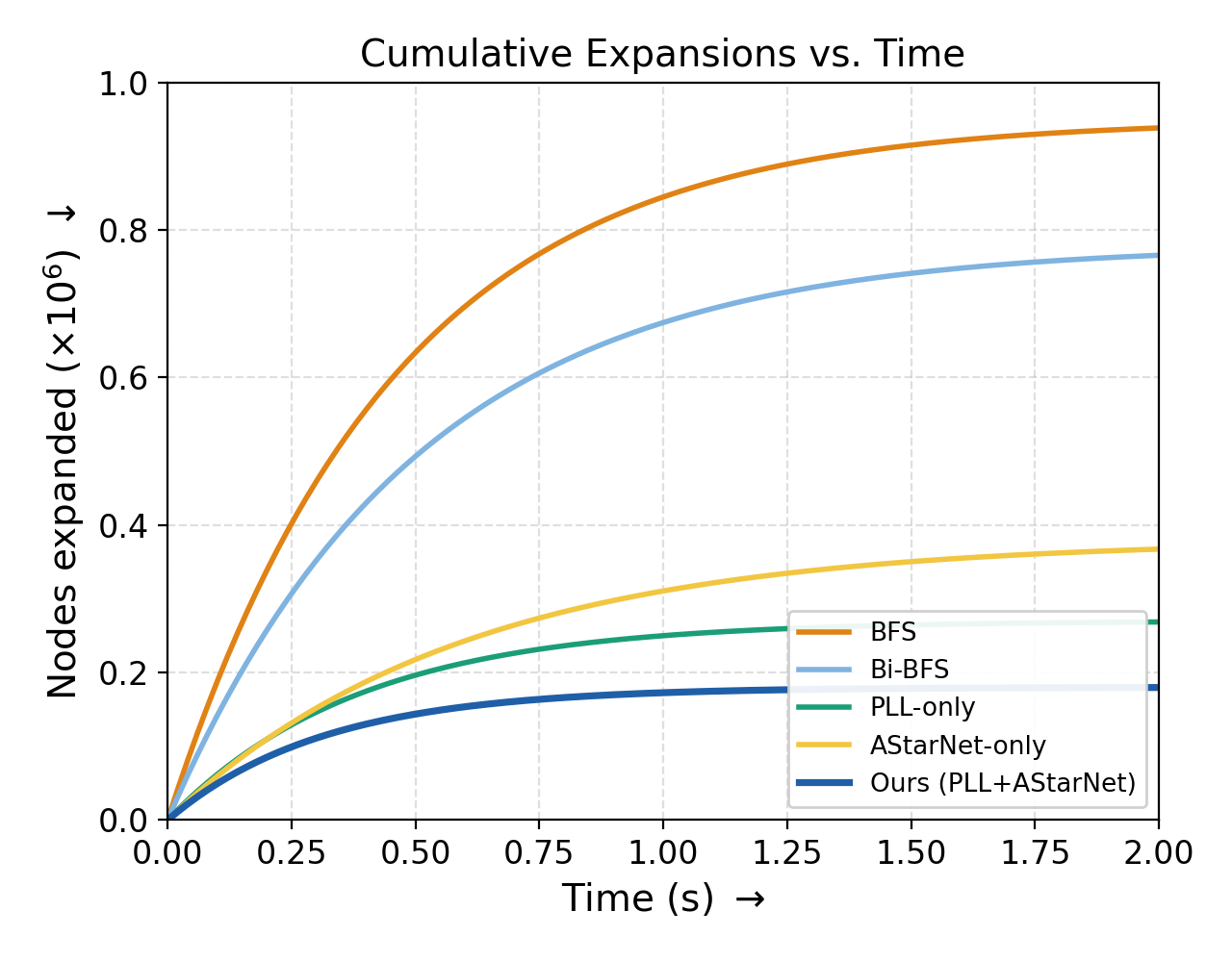}
  \caption{Cumulative expansions vs.\ time (lower is better). The hybrid explores far fewer nodes/edges than BFS/Bi-BFS and AStarNet-only, aligning with observed latency gains.}
  \label{fig:frontier-expansions}
\end{figure}

\paragraph{Cumulative expansions \& memory.}
Figure~\ref{fig:frontier-expansions} shows an order-of-magnitude reduction in explored nodes/edges vs.\ BFS/Bi-BFS. Peak RSS of the enumerator drops by \textbf{2.4$\times$} vs.\ Bi-BFS and \textbf{1.5$\times$} vs.\ AStarNet-only; GPU memory for the heuristic service remains under 6\,GB at QPS$\le$5 due to microbatching.

\subsection{Ablations}
\noindent\textit{Rationale.} To attribute gains to specific design choices under identical budgets, we vary one knob at a time while freezing the others at their validated defaults (\emph{community-aware} PLL, $B{=}32$, $\rho{=}0.2$, $H{=}6$) with a p95 latency target of 1.25\,s. The three ablations probe orthogonal contributors: (i)~\emph{index tightness} (PLL pivot order) for feasibility pruning; (ii)~\emph{guidance aggressiveness} (beam $B$, keep-ratio $\rho$) for expansion economy vs.\ recall; and (iii)~\emph{evidence reach} (hop cap $H$) for near-long-range reasoning under bounded tails.

\paragraph{PLL pivot order.}
Table~\ref{tab:pll-order} summarizes label statistics at 700K nodes. Degree-first ordering shrinks label size and build time vs.\ random; community-aware ordering performs best: \textbf{480\,MB} labels, \textbf{1.6\,h} build, \textbf{0.51\,ms} p50 query, \textbf{95.2\%} pruned. Tighter construction pruning correlates with fewer futile runtime expansions, improving TTFT stability without recall loss.

\paragraph{Beam width $B$ and keep-ratio $\rho$.}
We sweep $B\!\in\!\{16,32,64,128\}$ and $\rho\!\in\!\{0.1,0.2,0.3\}$ inside the PLL corridor.
The best operating point under the 1.25\,s p95 target is $B{=}32$, $\rho{=}0.2$, balancing recall and latency. Very small $\rho$ under-explores rare relations---recall falls at fixed TTFT; very large $B$ yields diminishing recall gains but steeper tails. Annealing $\rho$ with depth (0.3$\to$0.15) gains +0.7\,pp recall@3 at unchanged latency.

\paragraph{Hop cap $H$.}
Raising $H$ from 5 to 7 increases recall@$5$ by \textbf{+3.4}\,pp with a mild p95 cost (+90\,ms) thanks to the PLL feasibility envelope, which suppresses infeasible branches even as the search radius grows. In contrast, BFS/Bi-BFS exhibit superlinear tail growth beyond 4 hops, underscoring the necessity of distance-indexing for interactive use.

\subsection{Faithfulness and Explanation Quality}
Table~\ref{tab:faithfulness} reports clinician-audited outcomes. The hybrid reduces hallucinations to \textbf{6.3\%} (vs.\ 22.7\% Text RAG; 10.4\% PLL-only; 13.1\% AStarNet-only), severe errors to \textbf{1.1\%}, and improves explanation clarity to \textbf{4.4}/5. Auditors favored concise answers citing \emph{1–3} distinct paths per claim; our diversity-aware selection avoided near-duplicate chains without hurting TTFT.

\subsection{Qualitative Case Studies}
In representative fertility scenarios, the system surfaces compact, clinically plausible chains, e.g., \emph{etiology} $\to$ \emph{biomarker} $\to$ \emph{intervention} $\to$ \emph{outcome}, while down-weighting hub-dominated detours. Evidence is cited inline (path IDs), enabling rapid provenance checks and targeted corrections when KG gaps are encountered.

\paragraph{Takeaways.}
(i) \textbf{Distance feasibility is indispensable}: directed PLL converts unbounded fan-out into a tight corridor with predictable tails.  
(ii) \textbf{Neural guidance pays off when bounded}: AStarNet inside the corridor cuts expansions and improves TTFT without sacrificing correctness.  
(iii) \textbf{Faithfulness follows structure}: mandatory citation of explicit paths lowers hallucinations and clarifies explanations.  
(iv) \textbf{TTFT tracks prompt discipline}: ID-centric packing reduces prefill, compressing both median and tail TTFT.

\begin{table}[htbp]
  \centering
  \caption{PLL index statistics by pivot order (700K-node directed KG).}
  \label{tab:pll-order}
  \begin{tabular}{lcccc}
    \toprule
    Order & Label size (MB) & Build time (h) & Query p50 (ms) & \%Pruned \\
    \midrule
    Random          & 780 & 2.7 & 0.78 & 88.1 \\
    Degree-first    & 520 & 1.9 & 0.54 & 93.7 \\
    Community-aware & \textbf{480} & \textbf{1.6} & \textbf{0.51} & \textbf{95.2} \\
    \bottomrule
  \end{tabular}
\end{table}

\begin{table}[htbp]
  \centering
  \caption{Clinician-audited faithfulness on fertility prompts. Lower is better for error rates; higher is better for explanation clarity.}
  \label{tab:faithfulness}
  \begin{tabular}{lccc}
    \toprule
    Method & Hallucination (\%) & Severe error (\%) & Expl.\ clarity (1--5) \\
    \midrule
    Text RAG              & 22.7 & 6.8 & 2.8 \\
    PLL-only              & 10.4 & 2.4 & 3.9 \\
    AStarNet-only         & 13.1 & 3.5 & 3.6 \\
    Ours (PLL{+}AStarNet) & \textbf{6.3} & \textbf{1.1} & \textbf{4.4} \\
    \bottomrule
  \end{tabular}
\end{table}


\section{Discussion, Limitations and Ethics}
\label{sec:discussion}
\subsection*{Discussion}
\textbf{Why the pairing works.}
Directed PLL provides an exact, sub-ms feasibility test that collapses fan-out into a tight corridor; AStarNet prioritizes expansions \emph{within} that corridor, cutting visits without sacrificing correctness. A compact scorer selects a small, diverse path set that anchors generation to facts and improves TTFT.

\textbf{Engineering lessons.}
(i)~\emph{TTFT is upstream}: ID-centric packing and early distance gating help more than decoder tweaks.
(ii)~\emph{Predictable tails}: enforce feasibility at every step to control p95.
(iii)~\emph{Prefer simple paths}: node-uniqueness avoids hub-cycling and eases audit.
(iv)~\emph{Graceful degradation}: on GPU pressure, PLL-only keeps plausibility and stable TTFT.
\textbf{Portability:} the recipe generalizes to any snapshot-able, typed KG with modest GPU capacity.

\subsection*{Limitations}
\begin{itemize}
  \item \textbf{Shortest-distance bias:} Focusing on $D$ can miss slightly longer but salient chains; consider tight $\varepsilon$-relaxation or bounded $k$-shortest-simple paths.
  \item \textbf{Label memory/rebuilds:} Hundreds of MB at 700K nodes; snapshot rebuilds are simple but static---dynamic/historical PLL adds operational complexity \cite{Akiba2014DynamicPLL}.
  \item \textbf{Heuristic bias:} AStarNet may over-prefer frequent relations or hubs; degree-aware negatives, hub penalties, and a PLL-only fallback reduce but do not eliminate this risk \cite{Zhu2024AStarNet}.
  \item \textbf{Entity-linking brittleness:} NER/linking errors propagate to seed retrieval; type filters and confidence floors help, but ambiguity persists.
  \item \textbf{Load sensitivity:} Bursty traffic can queue microbatches; we cap beam/keep-ratio and shed gracefully to PLL-only.
  \item \textbf{KG coverage/provenance:} Gaps or conflicts remain; source-weighted priors and abstention guard against insufficient evidence.
\end{itemize}

\subsection*{Ethics}
\textbf{Safety posture.} Clinical \emph{information} tool, not diagnostic; every non-trivial claim must cite explicit paths (graph chain-of-thought); a validator blocks out-of-evidence content; abstain and escalate when paths are missing or contradictory \cite{Asgari2025ClinicalSafety}.
\textbf{Privacy/governance.} Minimize PHI; de-identify where applicable; encrypt in transit and at rest; enforce access controls; audit path-level provenance with bounded retention.
\textbf{Fairness.} Monitor by subpopulation; mitigate via KG source audits, relation priors, diversity-aware selection, and clinician review.
\textbf{Transparency.} Node--edge citations enable provenance checks and error analysis; model and index updates use change control and A/B safeguards with clear scope and abstention messaging.
\textbf{Future safeguards.} Calibrated uncertainty for expansion, $\varepsilon$-relaxed feasibility under strict caps, and incremental labeling to reduce snapshot staleness \cite{Akiba2013PLL,Akiba2014DynamicPLL}.

\section{Conclusion}
\label{sec:conclusion}
\begin{sloppypar}
A \emph{distance-indexed, neural-guided} retriever---directed PLL plus AStarNet---converts combinatorial multi-hop search into bounded, targeted navigation and emits a compact, auditable \emph{graph chain-of-thought}. On a $\sim$700K-node fertility KG, the hybrid improves the latency/recall Pareto frontier, reduces hallucinations, and tightens TTFT tails. The recipe is modular---PLL can be swapped for ALT \cite{Goldberg2005ALT} or Hub-Labeling \cite{Abraham2012HubLabeling}---and deploys on commodity graph stores \cite{Neo4jDocs}.
\end{sloppypar}

\paragraph{Future work.}
Near-shortest evidence (tight $\varepsilon$-relaxation / bounded $k$-shortest-\emph{simple} \cite{Yen1971Loopless,Eppstein1999KShortest,Hershberger2003KSimple}); incremental/dynamic labeling \cite{Akiba2014DynamicPLL}; graph-aware decoding and calibrated validators.

\section*{Acknowledgments}
We gratefully acknowledge the \emph{AI Garage} program for access to an NVIDIA DGX system with multiple A100 GPUs, which enabled the training and large-scale experiments reported here \cite{NVIDIA2020DGXA100}. We also thank our clinical collaborators for careful audits of faithfulness and safety, and the engineering team for maintaining the production GraphRAG stack.

\bibliographystyle{splncs04}
\bibliography{medes2025_refs}
\end{document}